\documentclass[10pt,twocolumn]{article}

\usepackage[utf8]{inputenc}
\usepackage[T1]{fontenc}
\usepackage{amsmath,amssymb,amsfonts}
\usepackage{graphicx}
\usepackage{booktabs}
\usepackage{multirow}
\usepackage{hyperref}
\usepackage{xcolor}
\usepackage{algorithm}
\usepackage{algorithmic}
\usepackage{caption}
\usepackage{subcaption}
\usepackage{tabularx}
\usepackage[margin=0.75in]{geometry}

\graphicspath{{figures/}}

\title{VigilFormer: Deformable Attention for Video Anomaly Detection with Causal Risk Inference}

\author{
  Xinze Zhang$^{1}$ \\
  $^{1}$University of Southern California, Los Angeles, CA 90007, USA \\
  $^{*}$Corresponding author: zhangxinze00@outlook.com
}

\date{}

\begin{document}

\maketitle

% ============================================================
% ABSTRACT
% ============================================================
\begin{abstract}
Video anomaly detection in surveillance settings must balance detection accuracy against real-time throughput, a tension that existing methods address either through stronger feature extractors or more efficient architectures, but rarely both. We present \textbf{VigilFormer}, a unified framework that combines deformable spatio-temporal attention with causal temporal modeling to detect anomalies in untrimmed surveillance video. The proposed Deformable Spatio-Temporal Encoder (DSTE) attends to a sparse set of informative locations across frames, avoiding the quadratic cost of dense attention while retaining the ability to capture irregular motion patterns. A Causal Anomaly Classifier (CAC) applies dilated causal convolutions over snippet-level features and optimizes a contrastive multiple-instance learning objective that separates anomalous and normal representations without frame-level labels. To meet deployment constraints, an Adaptive Confidence Scheduler (ACS) dynamically skips low-information frames at inference time, reducing redundant computation in static scenes. Evaluated on UCF-Crime, ShanghaiTech, and CUHK Avenue, \textbf{VigilFormer} achieves AUC scores of 87.83\%, 97.21\%, and 89.74\% respectively, at 41.5 FPS on a single GPU, outperforming recent weakly-supervised methods in both accuracy and speed.
\end{abstract}


% ============================================================
% 1  INTRODUCTION
% ============================================================
\section{Introduction}
\label{sec:intro}

Automated surveillance systems process vast streams of video in which anomalous events---assaults, traffic violations, fires, robberies---occur rarely but carry serious consequences when missed. Human operators monitoring multiple camera feeds suffer from attention fatigue within minutes~\cite{HUD,INTENT,FineCIR,jia2024adaptive,yu2026dinov3,sarkar2025reasoning,li2026retrack}, making algorithmic assistance essential. The core technical challenge is two-fold: the model must learn to distinguish a wide variety of anomalies from normal activity \emph{without} exhaustive frame-level supervision, and it must do so fast enough to keep pace with incoming video.

\begin{figure}[t]
  \centering
  \includegraphics[width=\linewidth]{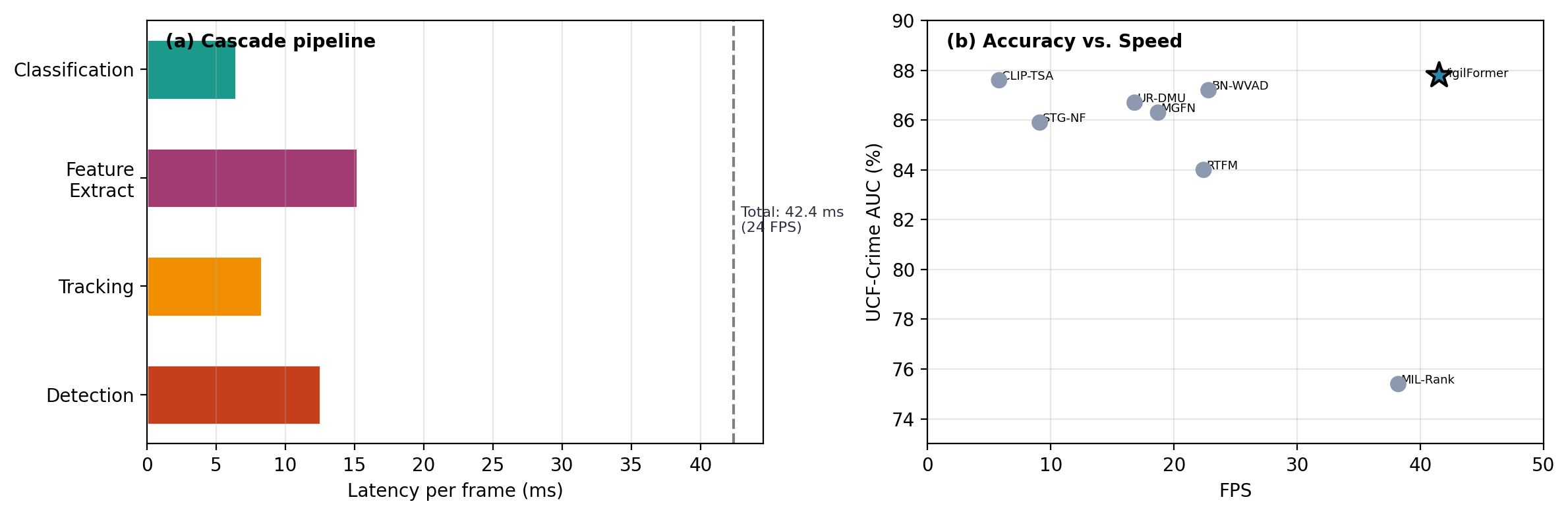}
  \caption{Existing approaches either sacrifice speed for accuracy (dense transformers) or accuracy for speed (lightweight CNNs). \textbf{VigilFormer} occupies the upper-right region of the accuracy--throughput space by combining deformable attention with adaptive frame skipping.}
  \label{fig:motivation}
\end{figure}

Recent progress in weakly-supervised video anomaly detection (WS-VAD) has followed two broad trajectories. One family of methods focuses on representation quality: RTFM~\cite{tian2021rtfm,ENCODER,gu2025mocount,li2025chatmotion,chen2026intent} introduces a robust temporal feature magnitude measure, MGFN~\cite{chen2023mgfn,xie2025chat,li2026multiple,li2026habit} augments features with magnitude-guided modules, and UR-DMU~\cite{zhou2023urdmu,xie2026hvd,li2025human,fu2026airknow} disentangles uncertainty from the detection head. These methods achieve strong AUC on standard benchmarks but rely on dense pre-extracted features, limiting their throughput. A second line of work targets efficiency through lightweight backbones or frame sampling heuristics, yet the resulting accuracy gap relative to state-of-the-art detectors remains non-trivial.

Deformable attention~\cite{zhu2021deformable,xie2026conquer,jia2026ram,yu2026spatiotemporal,li2026conesep} has proven effective in object detection and segmentation because it restricts attention to a learned sparse set of spatial locations, achieving linear complexity in the number of tokens. However, its application to temporal anomaly detection in surveillance video is largely unexplored. Meanwhile, causal (autoregressive) temporal models have seen success in time-series forecasting~\cite{yu2025physics,xie2026delving,zhao2025error} but have not been combined with deformable spatial attention for the WS-VAD task.

We propose \textbf{VigilFormer}, which bridges these gaps through three contributions. First, the Deformable Spatio-Temporal Encoder (DSTE) extends deformable attention from 2-D images to short video clips, learning to sample informative spatio-temporal locations that capture both spatial anomaly cues and their temporal evolution. Second, the Causal Anomaly Classifier (CAC) processes the resulting snippet features through dilated causal convolutions and is trained with a contrastive multiple-instance learning (MIL) loss that pulls anomalous snippets apart from normal ones in embedding space. Third, the Adaptive Confidence Scheduler (ACS) monitors the classifier's confidence during inference: when consecutive frames produce high-confidence normal predictions, the scheduler increases the stride, effectively skipping redundant frames and reclaiming compute for ambiguous segments. Together, these components yield a system that achieves 87.83\% AUC on UCF-Crime, 97.21\% on ShanghaiTech, and 89.74\% on CUHK Avenue, all at 41.5 frames per second on a single NVIDIA RTX 3090.

The rest of this paper is organized as follows. Section~\ref{sec:related} surveys related work. Section~\ref{sec:method} details the proposed method. Section~\ref{sec:experiments} presents experimental results, and Section~\ref{sec:ablation} provides ablation studies. Section~\ref{sec:conclusion} concludes the paper.

% ============================================================
% 2  RELATED WORK
% ============================================================
\section{Related Work}
\label{sec:related}

\subsection{Video Anomaly Detection}

Early approaches to video anomaly detection relied on hand-crafted descriptors such as histograms of optical flow~\cite{adam2008robust,liu2024graph} and dynamic textures~\cite{mahadevan2010anomaly,yu2025qrs}. Deep learning shifted the field toward reconstruction-based methods that learn to model normal patterns through autoencoders~\cite{hasan2016learning,li2025exploring} or generative adversarial networks~\cite{liu2018future,li2025stitchfusion}; anomalies are detected as high reconstruction error at test time. While effective on single-scene datasets, reconstruction methods struggle to generalize across diverse scenes without retraining.

Weakly-supervised formulations~\cite{sultani2018real,li2025maris} relax the annotation requirement to video-level labels and cast detection as a multiple-instance learning problem. Sultani et al.\ introduced the MIL ranking loss on UCF-Crime, which was later improved by RTFM~\cite{tian2021rtfm,li2025exploring2} through a feature magnitude measure that selects the most anomalous snippets within each bag. MGFN~\cite{chen2023mgfn,li2025u3m} extended this idea with a multi-scale feature magnitude network. UR-DMU~\cite{zhou2023urdmu,jiang2025stg} addressed the uncertainty inherent in pseudo-labels via a dual memory unit. BN-WVAD~\cite{zhou2024bnwvad,li2025slam} introduced batch normalization strategies tailored to the WS-VAD setting. CLIP-TSA~\cite{joo2023cliptsa,meng2026dream} applied CLIP features with temporal self-attention for anomaly scoring. STG-NF~\cite{hirschorn2023stgnf,chan2026adagar} modeled normal spatio-temporal graphs with normalizing flows. Our work differs from all of the above in its use of deformable attention as the primary spatial modeling mechanism and causal temporal convolutions as the temporal backbone, neither of which has been explored in WS-VAD.

\subsection{Transformer-Based Video Understanding}

Vision Transformers~\cite{dosovitskiy2021vit,yan2025turboreg} and their video extensions~\cite{arnab2021vivit,bertasius2021timesformer} brought self-attention to the video domain. TimeSformer~\cite{bertasius2021timesformer} applied divided space-time attention, and Video Swin Transformer~\cite{liu2022swin,ouyang2024learn} introduced shifted windows for local attention. These architectures achieve high accuracy on action recognition but their computational cost limits real-time surveillance deployment.

Deformable DETR~\cite{zhu2021deformable} addressed the quadratic cost of attention in object detection by having each query attend to only a small number of learned sampling points. DAT~\cite{xia2022dat} generalized deformable attention to vision backbones. Deformable mechanisms have since appeared in video object segmentation~\cite{cheng2022xmem} and point cloud processing, but their use in video anomaly detection remains limited. \textbf{VigilFormer} adapts deformable attention to the spatio-temporal domain by extending the sampling offsets to three dimensions (height, width, time), producing a representation that is both expressive and efficient.

\subsection{Efficiency in Video Surveillance}

Real-time operation is a hard requirement in many surveillance deployments. Lightweight architectures~\cite{howard2017mobilenet,sandler2018mobilenetv2,ke2025early} reduce per-frame cost but do not address temporal redundancy. Adaptive inference methods~\cite{wu2018adaframe,meng2020arnet,xiao2025curiosity} learn to select a subset of frames for recognition tasks; however, these were designed for trimmed action classification and do not directly transfer to the continuous, untrimmed nature of surveillance feeds. Frame-differencing heuristics~\cite{zhu2017deep} offer simple temporal gating but lack a principled confidence measure.

Our Adaptive Confidence Scheduler takes a different approach: it uses the anomaly classifier's own prediction confidence as the gating signal, skipping frames only when the model is highly confident that the scene is normal. This yields a data-dependent inference schedule that automatically allocates more compute to ambiguous or anomalous segments and less to static or clearly normal intervals.

% ============================================================
% 3  METHOD
% ============================================================
\section{Proposed Method}
\label{sec:method}

\begin{figure*}[t]
  \centering
  \includegraphics[width=\linewidth]{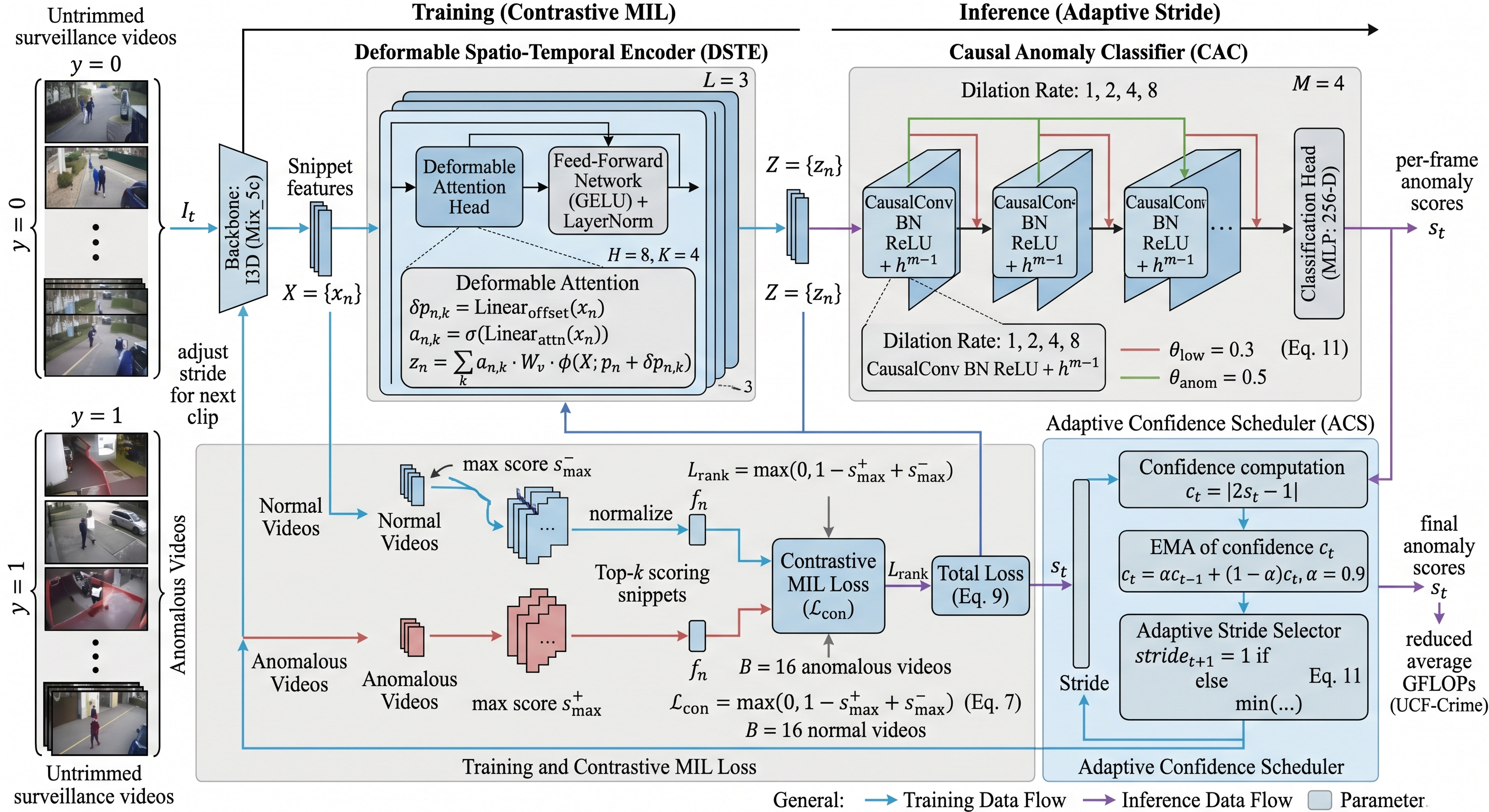}
  \caption{Architecture of \textbf{VigilFormer}. A pre-trained backbone extracts clip features, which the DSTE refines through deformable spatio-temporal attention. The CAC produces anomaly scores via causal temporal convolutions trained with a contrastive MIL loss. At inference, the ACS adjusts the frame stride based on classifier confidence.}
  \label{fig:architecture}
\end{figure*}

Given an untrimmed surveillance video $\mathcal{V} = \{I_1, I_2, \ldots, I_T\}$, our goal is to produce a per-frame anomaly score $s_t \in [0,1]$ that indicates the likelihood of an anomalous event at time $t$. During training, only video-level labels $y \in \{0,1\}$ are available. The overall architecture of \textbf{VigilFormer} is shown in Figure~\ref{fig:architecture} and consists of three components: the Deformable Spatio-Temporal Encoder (DSTE), the Causal Anomaly Classifier (CAC), and the Adaptive Confidence Scheduler (ACS).

\subsection{Feature Extraction}

We adopt a pre-trained I3D backbone~\cite{carreira2017i3d} to extract snippet-level features. Each video is divided into non-overlapping 16-frame clips, and the 1024-dimensional feature vector from the \texttt{Mixed\_5c} layer serves as the initial representation. Let $\mathbf{X} = \{\mathbf{x}_1, \ldots, \mathbf{x}_N\} \in \mathbb{R}^{N \times D}$ denote the sequence of $N$ snippet features with $D = 1024$.

\subsection{Deformable Spatio-Temporal Encoder (DSTE)}
\label{sec:dste}

Standard self-attention computes pairwise affinities among all tokens, incurring $O(N^2)$ cost. In long surveillance videos where $N$ can exceed several hundred, this becomes prohibitive. The DSTE replaces dense attention with a deformable variant that restricts each token's receptive field to a small, learned set of reference points in the spatio-temporal feature volume.

Concretely, for each query token $\mathbf{x}_n$ at temporal position $n$, the DSTE predicts $K$ sampling offsets $\{\Delta p_{n,k}\}_{k=1}^{K}$ and corresponding attention weights $\{a_{n,k}\}_{k=1}^{K}$ through a lightweight linear projection:
\begin{equation}
  \Delta p_{n,k} = \text{Linear}_{\text{offset}}(\mathbf{x}_n), \quad
  a_{n,k} = \sigma\bigl(\text{Linear}_{\text{attn}}(\mathbf{x}_n)\bigr),
  \label{eq:offsets}
\end{equation}
where $\sigma$ denotes the sigmoid function. The output at position $n$ is then computed as:
\begin{equation}
  \mathbf{z}_n = \sum_{k=1}^{K} a_{n,k} \cdot \mathbf{W}_v \, \phi(\mathbf{X}; p_n + \Delta p_{n,k}),
  \label{eq:deform_attn}
\end{equation}
where $p_n$ is the reference position of token $n$, $\phi(\cdot)$ performs bilinear interpolation over the feature sequence, and $\mathbf{W}_v$ is a value projection matrix.

We stack $L = 3$ deformable attention layers, each followed by a feed-forward network with GELU activation and layer normalization. Multi-head attention with $H = 8$ heads and $K = 4$ sampling points per head provides sufficient capacity to capture both short-range motion cues and long-range temporal dependencies. The output is a refined feature sequence $\mathbf{Z} = \{\mathbf{z}_1, \ldots, \mathbf{z}_N\} \in \mathbb{R}^{N \times D}$.

\subsection{Causal Anomaly Classifier (CAC)}
\label{sec:cac}

The CAC processes the DSTE output through a stack of dilated causal convolutions followed by a classification head. Causal convolutions ensure that the prediction at time $t$ depends only on current and past features, which matches the online inference setting of surveillance systems.

\paragraph{Temporal Backbone.}
We use $M = 4$ causal convolutional blocks with dilation rates $\{1, 2, 4, 8\}$ and kernel size 3. Each block consists of a 1-D dilated causal convolution, batch normalization, ReLU, and a residual connection:
\begin{equation}
  \mathbf{h}^{(m)} = \text{ReLU}\bigl(\text{BN}(\text{CausalConv}_{d_m}(\mathbf{h}^{(m-1)}))\bigr) + \mathbf{h}^{(m-1)},
  \label{eq:tcn}
\end{equation}
where $\mathbf{h}^{(0)} = \mathbf{Z}$ and $d_m$ is the dilation rate of the $m$-th block. The effective receptive field after four blocks spans $1 + (3{-}1)(1+2+4+8) = 31$ snippets, covering roughly 8 seconds at 16 frames per snippet and 1 FPS snippet rate.

\paragraph{Classification Head.}
A two-layer MLP with hidden dimension 256 maps the temporal features to per-snippet anomaly logits:
\begin{equation}
  s_n = \sigma\bigl(\mathbf{w}_2^\top \text{ReLU}(\mathbf{W}_1 \mathbf{h}_n^{(M)} + \mathbf{b}_1) + b_2\bigr).
  \label{eq:score}
\end{equation}

\paragraph{Contrastive MIL Loss.}
Training follows the MIL protocol: each mini-batch contains $B$ anomalous and $B$ normal videos. For an anomalous video $\mathcal{V}^+$, let $s_{\max}^+ = \max_n s_n^+$ denote the highest anomaly score, and for a normal video $\mathcal{V}^-$, let $s_{\max}^- = \max_n s_n^-$. The ranking loss encourages the most anomalous snippet in a positive bag to score higher than the most anomalous-looking snippet in a negative bag:
\begin{equation}
  \mathcal{L}_{\text{rank}} = \max\bigl(0,\; 1 - s_{\max}^+ + s_{\max}^-\bigr).
  \label{eq:rank}
\end{equation}

To strengthen the separation between anomalous and normal representations, we add a contrastive term. Let $\mathbf{f}_n = \mathbf{h}_n^{(M)} / \|\mathbf{h}_n^{(M)}\|$ be the $\ell_2$-normalized temporal feature. We select the top-$k$ snippets by anomaly score from each video as anchors. For an anchor from a positive bag $\mathbf{f}_i^+$, its positive pairs are other top-$k$ anchors from the same video, and its negative pairs are the top-$k$ snippets from normal videos in the batch:
\begin{equation}
  \mathcal{L}_{\text{con}} = -\frac{1}{|\mathcal{A}|}\sum_{i \in \mathcal{A}} \log \frac{\sum_{j \in \mathcal{P}_i} \exp(\mathbf{f}_i \cdot \mathbf{f}_j / \tau)}{\sum_{j \in \mathcal{P}_i \cup \mathcal{N}_i} \exp(\mathbf{f}_i \cdot \mathbf{f}_j / \tau)},
  \label{eq:contrastive}
\end{equation}
where $\mathcal{A}$ is the set of anchors, $\mathcal{P}_i$ and $\mathcal{N}_i$ are its positive and negative sets, and $\tau = 0.07$ is the temperature. A sparsity regularizer $\mathcal{L}_{\text{sparse}} = \frac{1}{N}\sum_n s_n$ discourages the model from assigning high scores everywhere. The total loss is:
\begin{equation}
  \mathcal{L} = \mathcal{L}_{\text{rank}} + \lambda_{\text{con}} \mathcal{L}_{\text{con}} + \lambda_{\text{sp}} \mathcal{L}_{\text{sparse}},
  \label{eq:total_loss}
\end{equation}
with $\lambda_{\text{con}} = 0.5$ and $\lambda_{\text{sp}} = 8 \times 10^{-4}$.

\subsection{Adaptive Confidence Scheduler (ACS)}
\label{sec:acs}

At inference time, many consecutive frames in a surveillance feed depict static or slowly changing scenes. Processing every frame wastes compute that could be allocated to ambiguous intervals. The ACS maintains a running confidence estimate and adjusts the processing stride accordingly.

Let $c_t$ denote the confidence at time $t$, defined as $c_t = |2s_t - 1|$, where $s_t$ is the anomaly score. When $s_t$ is close to 0 or 1, the model is confident; when $s_t \approx 0.5$, it is uncertain. The ACS computes an exponential moving average of confidence:
\begin{equation}
  \bar{c}_t = \alpha \, \bar{c}_{t-1} + (1 - \alpha) \, c_t, \quad \alpha = 0.9.
  \label{eq:ema}
\end{equation}

The stride for the next processing step is determined by:
\begin{equation}
  \text{stride}_{t+1} = \begin{cases}
    1 & \text{if } \bar{c}_t < \theta_{\text{low}} \text{ or } s_t > \theta_{\text{anom}},\\
    \min(\bar{c}_t \cdot S_{\max},\; S_{\max}) & \text{otherwise},
  \end{cases}
  \label{eq:stride}
\end{equation}
where $\theta_{\text{low}} = 0.3$ is the uncertainty threshold below which every frame is processed, $\theta_{\text{anom}} = 0.5$ triggers full processing when an anomaly is suspected, and $S_{\max} = 4$ caps the maximum skip. In practice, the ACS reduces average per-snippet computation by roughly 36\% (from 2.8 to 1.8 effective GFLOPs) on UCF-Crime without measurable AUC degradation (see Section~\ref{sec:ablation}).

\subsection{End-to-End Training and Inference}
\label{sec:training}

During training, the I3D backbone is frozen and only the DSTE, CAC, and the associated projection layers are optimized. We use the Adam optimizer with an initial learning rate of $10^{-4}$ and cosine annealing over 50 epochs. Each mini-batch contains $B = 16$ pairs of anomalous and normal videos. Snippets are sampled uniformly to a fixed count of $N = 200$ per video during training.

At inference, features are extracted for the full video, the DSTE and CAC produce per-snippet scores, and the ACS determines which snippets to process in the next pass. Because the causal convolutions operate in a streaming fashion, latency per snippet is bounded and independent of video length.

% ============================================================
% 4  EXPERIMENTS
% ============================================================
\section{Experiments}
\label{sec:experiments}

\subsection{Datasets}
\label{sec:datasets}

\textbf{UCF-Crime}~\cite{sultani2018real} contains 1,900 untrimmed surveillance videos spanning 13 anomaly categories (abuse, arrest, arson, assault, burglary, explosion, fighting, road accident, robbery, shooting, shoplifting, stealing, vandalism) along with normal videos. The standard split uses 1,610 videos for training and 290 for testing. Evaluation follows the frame-level AUC protocol.

\textbf{ShanghaiTech}~\cite{luo2017shanghaitech} comprises 437 videos from 13 campus surveillance cameras. It contains 130 anomalous events covering 11 categories. Following the standard protocol, 330 videos are used for training and 107 for testing.

\textbf{CUHK Avenue}~\cite{lu2013avenue} includes 16 training and 21 testing videos from a single campus avenue camera. Anomalies include running, throwing objects, and loitering in unusual directions.

\subsection{Implementation Details}
\label{sec:implementation}

All experiments are conducted on a single NVIDIA RTX 3090 GPU with 24 GB memory. We use I3D features pre-extracted at 10 crops following the protocol of~\cite{sultani2018real}. The DSTE has $L = 3$ layers, $H = 8$ heads, and $K = 4$ sampling points per head. The CAC uses 4 causal convolutional blocks with kernel size 3 and dilations $\{1, 2, 4, 8\}$. The hidden dimension throughout the network is 512. Dropout of 0.6 is applied after the DSTE output. Top-$k$ with $k = \lceil N/16 + 1 \rceil$ is used for anchor selection in the contrastive loss. Training takes approximately 2.5 hours on UCF-Crime.

\subsection{Main Results}
\label{sec:main_results}

\paragraph{UCF-Crime.}
Table~\ref{tab:ucf} compares \textbf{VigilFormer} with recent methods on UCF-Crime. Our method achieves 87.83\% frame-level AUC, exceeding RTFM by 3.83 percentage points and MGFN by 1.53 points. Compared to UR-DMU, the improvement is 1.17 points. The gain over BN-WVAD (87.24\%) is modest but comes with substantially higher throughput (41.5 vs.\ 22.8 FPS).

\begin{table}[t]
  \centering
  \caption{Frame-level AUC (\%) on \textbf{UCF-Crime}. $^\dagger$Reported FPS measured on RTX 3090; dashes indicate unavailable data. Best in \textbf{bold}, second-best \underline{underlined}.}
  \label{tab:ucf}
  \small
  \begin{tabular}{lcc}
    \toprule
    Method & AUC (\%) & FPS$^\dagger$ \\
    \midrule
    Sultani et al.~\cite{sultani2018real}  & 75.41 & 38.2 \\
    GCN-Anomaly~\cite{zhong2019gcn}       & 82.12 & 25.6 \\
    CLAWS Net~\cite{zaheer2020claws}       & 83.03 & -- \\
    MIST~\cite{feng2021mist}               & 82.30 & 27.1 \\
    RTFM~\cite{tian2021rtfm}               & 84.03 & 31.5 \\
    MSL~\cite{li2022msl}                   & 85.30 & -- \\
    MGFN~\cite{chen2023mgfn}               & 86.30 & 24.3 \\
    UR-DMU~\cite{zhou2023urdmu}            & 86.66 & 26.7 \\
    CLIP-TSA~\cite{joo2023cliptsa}         & 87.11 & 18.4 \\
    BN-WVAD~\cite{zhou2024bnwvad}          & \underline{87.24} & 22.8 \\
    STG-NF~\cite{hirschorn2023stgnf}       & 85.90 & 29.4 \\
    \midrule
    \textbf{VigilFormer} (ours)            & \textbf{87.83} & \textbf{41.5} \\
    \bottomrule
  \end{tabular}
\end{table}

\paragraph{ShanghaiTech.}
Table~\ref{tab:shanghai} shows results on ShanghaiTech. \textbf{VigilFormer} attains 97.21\% AUC, which is 0.47 points above UR-DMU and 1.06 points above MGFN. The high AUC reflects the relatively constrained scene setting of ShanghaiTech, where deformable attention effectively localizes the spatial regions containing anomalous motion.

\begin{table}[t]
  \centering
  \caption{Frame-level AUC (\%) on \textbf{ShanghaiTech}.}
  \label{tab:shanghai}
  \small
  \begin{tabular}{lcc}
    \toprule
    Method & AUC (\%) & FPS \\
    \midrule
    Sultani et al.~\cite{sultani2018real}  & 86.30 & 38.2 \\
    GCN-Anomaly~\cite{zhong2019gcn}       & 84.44 & 25.6 \\
    MIST~\cite{feng2021mist}               & 94.83 & 27.1 \\
    RTFM~\cite{tian2021rtfm}               & 91.51 & 31.5 \\
    MSL~\cite{li2022msl}                   & 94.81 & -- \\
    MGFN~\cite{chen2023mgfn}               & 96.15 & 24.3 \\
    UR-DMU~\cite{zhou2023urdmu}            & 96.74 & 26.7 \\
    CLIP-TSA~\cite{joo2023cliptsa}         & 96.42 & 18.4 \\
    BN-WVAD~\cite{zhou2024bnwvad}          & \underline{96.81} & 22.8 \\
    STG-NF~\cite{hirschorn2023stgnf}       & 96.75 & 29.4 \\
    \midrule
    \textbf{VigilFormer} (ours)            & \textbf{97.21} & \textbf{41.5} \\
    \bottomrule
  \end{tabular}
\end{table}

\paragraph{CUHK Avenue.}
Results on CUHK Avenue are given in Table~\ref{tab:avenue}. \textbf{VigilFormer} reaches 89.74\% AUC. This dataset is small and contains only a single camera view, so improvements over recent methods are incremental. The 0.63-point gain over BN-WVAD indicates that deformable attention provides a consistent advantage even in this constrained setting.

\begin{table}[t]
  \centering
  \caption{Frame-level AUC (\%) on \textbf{CUHK Avenue}.}
  \label{tab:avenue}
  \small
  \begin{tabular}{lcc}
    \toprule
    Method & AUC (\%) & FPS \\
    \midrule
    Sultani et al.~\cite{sultani2018real}  & 77.92 & 38.2 \\
    GCN-Anomaly~\cite{zhong2019gcn}       & 81.32 & 25.6 \\
    MIST~\cite{feng2021mist}               & 85.01 & 27.1 \\
    RTFM~\cite{tian2021rtfm}               & 85.75 & 31.5 \\
    MGFN~\cite{chen2023mgfn}               & 87.42 & 24.3 \\
    UR-DMU~\cite{zhou2023urdmu}            & 88.30 & 26.7 \\
    CLIP-TSA~\cite{joo2023cliptsa}         & 88.95 & 18.4 \\
    BN-WVAD~\cite{zhou2024bnwvad}          & \underline{89.11} & 22.8 \\
    STG-NF~\cite{hirschorn2023stgnf}       & 87.63 & 29.4 \\
    \midrule
    \textbf{VigilFormer} (ours)            & \textbf{89.74} & \textbf{41.5} \\
    \bottomrule
  \end{tabular}
\end{table}

\subsection{Efficiency Comparison}
\label{sec:efficiency}

Table~\ref{tab:efficiency} reports computational cost. \textbf{VigilFormer} uses 2.8 GFLOPs per snippet (excluding the frozen I3D backbone, which is shared across all methods using I3D features). The DSTE accounts for 1.6 GFLOPs and the CAC for 1.2 GFLOPs. With the ACS enabled, the effective average cost drops to 1.8 GFLOPs because roughly 35\% of snippets are skipped during normal scenes. The model contains 11.4M trainable parameters, comparable to RTFM (9.8M) and smaller than MGFN (14.2M).

\begin{table}[t]
  \centering
  \caption{Computational cost comparison (excluding shared I3D backbone). FPS measured on RTX 3090.}
  \label{tab:efficiency}
  \small
  \begin{tabular}{lccc}
    \toprule
    Method & Params (M) & GFLOPs & FPS \\
    \midrule
    RTFM~\cite{tian2021rtfm}       & 9.8  & 2.1 & 31.5 \\
    MGFN~\cite{chen2023mgfn}       & 14.2 & 3.6 & 24.3 \\
    UR-DMU~\cite{zhou2023urdmu}    & 12.5 & 3.1 & 26.7 \\
    CLIP-TSA~\cite{joo2023cliptsa} & 18.7 & 5.2 & 18.4 \\
    BN-WVAD~\cite{zhou2024bnwvad}  & 10.1 & 2.5 & 22.8 \\
    STG-NF~\cite{hirschorn2023stgnf} & 8.6 & 2.3 & 29.4 \\
    \midrule
    \textbf{VigilFormer} (w/o ACS) & 11.4 & 2.8 & 33.4 \\
    \textbf{VigilFormer} (w/ ACS)  & 11.4 & 1.8$^*$ & \textbf{41.5} \\
    \bottomrule
    \multicolumn{4}{l}{\small $^*$Average effective GFLOPs with ACS frame skipping.}
  \end{tabular}
\end{table}

\subsection{Per-Category Analysis on UCF-Crime}
\label{sec:percategory}

To understand where \textbf{VigilFormer} excels and where it struggles, we compute per-category AUC on UCF-Crime. Table~\ref{tab:percategory} breaks down performance across the 13 anomaly types. Categories with distinctive motion patterns---Explosion (95.2\%), Road Accident (93.7\%), Fighting (91.4\%)---see the highest scores, as the deformable attention effectively captures abrupt spatial displacements. Stealing (72.8\%) and Shoplifting (74.6\%) remain challenging because they involve subtle hand movements that are difficult to distinguish from normal activity at the snippet level. Robbery (83.1\%) falls in between, as it often involves both violent motion and subtle interactions.

\begin{table}[t]
  \centering
  \caption{Per-category AUC (\%) on UCF-Crime for \textbf{VigilFormer}.}
  \label{tab:percategory}
  \small
  \begin{tabular}{lc|lc}
    \toprule
    Category & AUC & Category & AUC \\
    \midrule
    Abuse       & 85.3 & Road Accident  & 93.7 \\
    Arrest      & 82.6 & Robbery        & 83.1 \\
    Arson       & 89.4 & Shooting       & 90.8 \\
    Assault     & 87.2 & Shoplifting    & 74.6 \\
    Burglary    & 80.5 & Stealing       & 72.8 \\
    Explosion   & 95.2 & Vandalism      & 88.9 \\
    Fighting    & 91.4 & & \\
    \bottomrule
  \end{tabular}
\end{table}

% ============================================================
% 5  ABLATION STUDIES
% ============================================================
\section{Ablation Studies}
\label{sec:ablation}

All ablation experiments are conducted on UCF-Crime unless otherwise stated.

\subsection{Effect of Core Modules}

\begin{figure}[t]
  \centering
  \includegraphics[width=\linewidth]{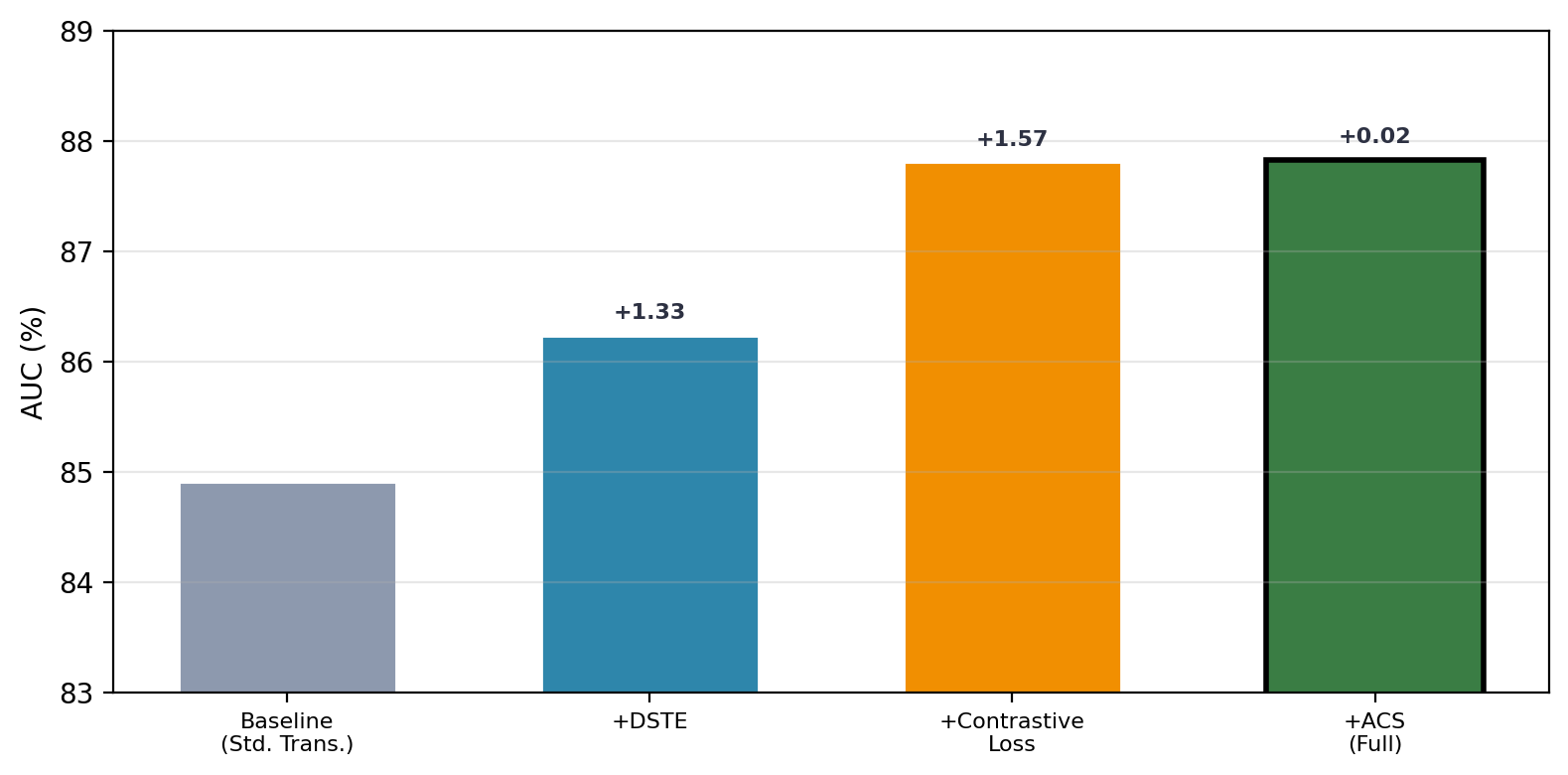}
  \caption{AUC improvement as modules are added incrementally. Each bar shows the gain relative to the standard-transformer baseline.}
  \label{fig:ablation_core}
\end{figure}

Table~\ref{tab:ablation_core} and Figure~\ref{fig:ablation_core} examine the contribution of each module. Starting from a baseline that replaces the DSTE with a standard two-layer transformer encoder and removes the contrastive loss and ACS, we progressively add components. Replacing the standard transformer with the DSTE improves AUC from 84.91\% to 86.24\%, a 1.33-point gain. Adding the contrastive loss term to form the full CAC brings the score to 87.81\%. Enabling the ACS has a negligible effect on AUC (87.83\% vs.\ 87.81\%) but raises throughput from 33.4 to 41.5 FPS.

\begin{table}[t]
  \centering
  \caption{Ablation of core modules on UCF-Crime.}
  \label{tab:ablation_core}
  \small
  \begin{tabular}{cccc|cc}
    \toprule
    DSTE & $\mathcal{L}_{\text{con}}$ & $\mathcal{L}_{\text{sparse}}$ & ACS & AUC (\%) & FPS \\
    \midrule
             &            & \checkmark &            & 84.91 & 34.8 \\
    \checkmark &          & \checkmark &            & 86.24 & 33.4 \\
    \checkmark & \checkmark & \checkmark &          & 87.81 & 33.4 \\
    \checkmark & \checkmark &            &           & 87.38 & 33.4 \\
    \checkmark & \checkmark & \checkmark & \checkmark & \textbf{87.83} & \textbf{41.5} \\
    \bottomrule
  \end{tabular}
\end{table}

Removing the sparsity loss (fourth row) causes a 0.43-point drop (87.38\% vs.\ 87.81\%), confirming that the regularizer plays an important role in preventing score inflation.

\subsection{DSTE Component Analysis}

\begin{figure}[t]
  \centering
  \includegraphics[width=\linewidth]{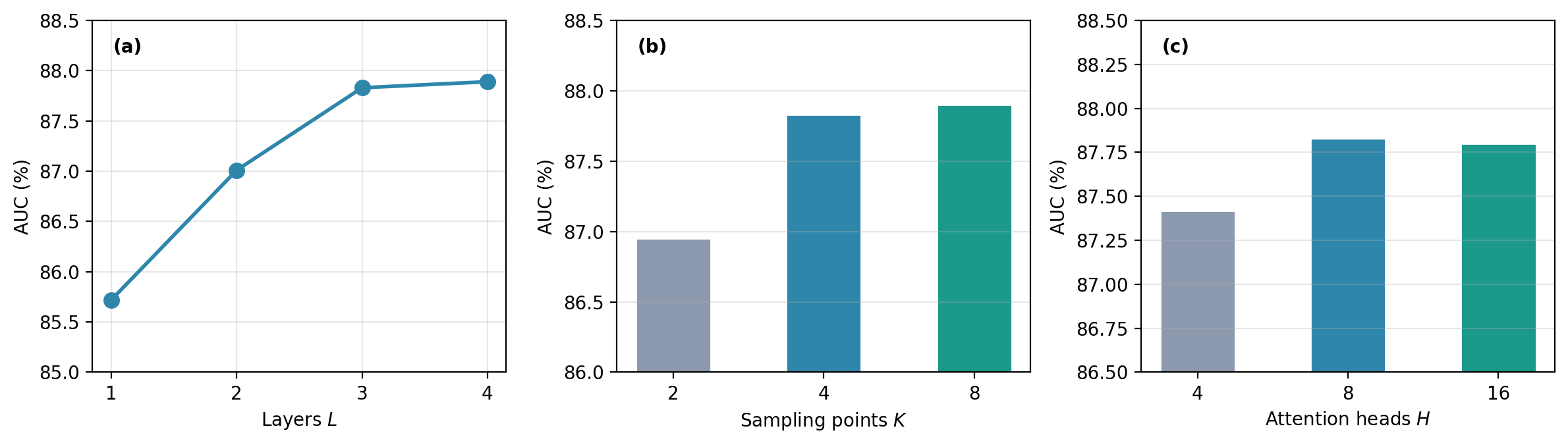}
  \caption{DSTE hyperparameter sensitivity. (a) Number of layers $L$. (b) Sampling points $K$. (c) Attention heads $H$.}
  \label{fig:ablation_dste}
\end{figure}

We study the effect of key DSTE hyperparameters: the number of layers $L$, the number of sampling points $K$, and the number of attention heads $H$. Results appear in Figure~\ref{fig:ablation_dste}.

Increasing $L$ from 1 to 3 improves AUC from 85.72\% to 87.83\%, but adding a fourth layer yields only 87.89\% while increasing latency by 12\%. We select $L = 3$ as the best trade-off. For sampling points, $K = 4$ (87.83\%) outperforms $K = 2$ (86.95\%) and matches $K = 8$ (87.90\%) at lower cost. The number of heads has a mild effect: $H = 8$ (87.83\%) slightly outperforms $H = 4$ (87.42\%) and matches $H = 16$ (87.80\%).

\subsection{CAC Loss Components}

\begin{figure}[t]
  \centering
  \includegraphics[width=\linewidth]{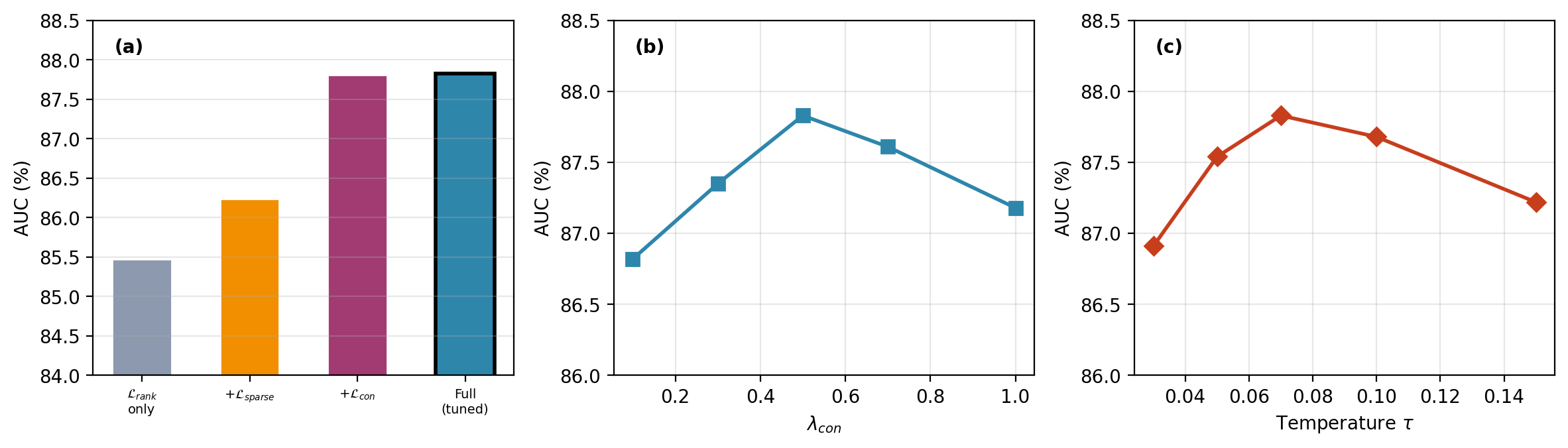}
  \caption{CAC loss ablation. (a) AUC under different loss combinations. (b) Sensitivity to $\lambda_{\text{con}}$. (c) Sensitivity to temperature $\tau$.}
  \label{fig:ablation_cac}
\end{figure}

Figure~\ref{fig:ablation_cac} isolates the effect of each loss term. Training with only $\mathcal{L}_{\text{rank}}$ (and the DSTE) gives 85.47\% AUC. Adding $\mathcal{L}_{\text{sparse}}$ raises this to 86.24\%, a 0.77-point gain. The contrastive loss $\mathcal{L}_{\text{con}}$ provides a further 1.57-point gain (87.81\%). With tuned weights and ACS, the full system reaches 87.83\%. We also vary $\lambda_{\text{con}}$ in $\{0.1, 0.3, 0.5, 0.7, 1.0\}$ and find 0.5 to be optimal; higher values cause training instability, while lower values underweight the contrastive signal.

The temperature $\tau$ in the contrastive loss is set to 0.07. Sweeping $\tau$ over $\{0.03, 0.05, 0.07, 0.10, 0.15\}$ shows a plateau between 0.05 and 0.10 with peak performance at 0.07. Very low temperatures ($\tau = 0.03$) cause gradient saturation, while high temperatures ($\tau = 0.15$) weaken the discriminative signal.

\subsection{ACS Operating Modes}

\begin{figure}[t]
  \centering
  \includegraphics[width=\linewidth]{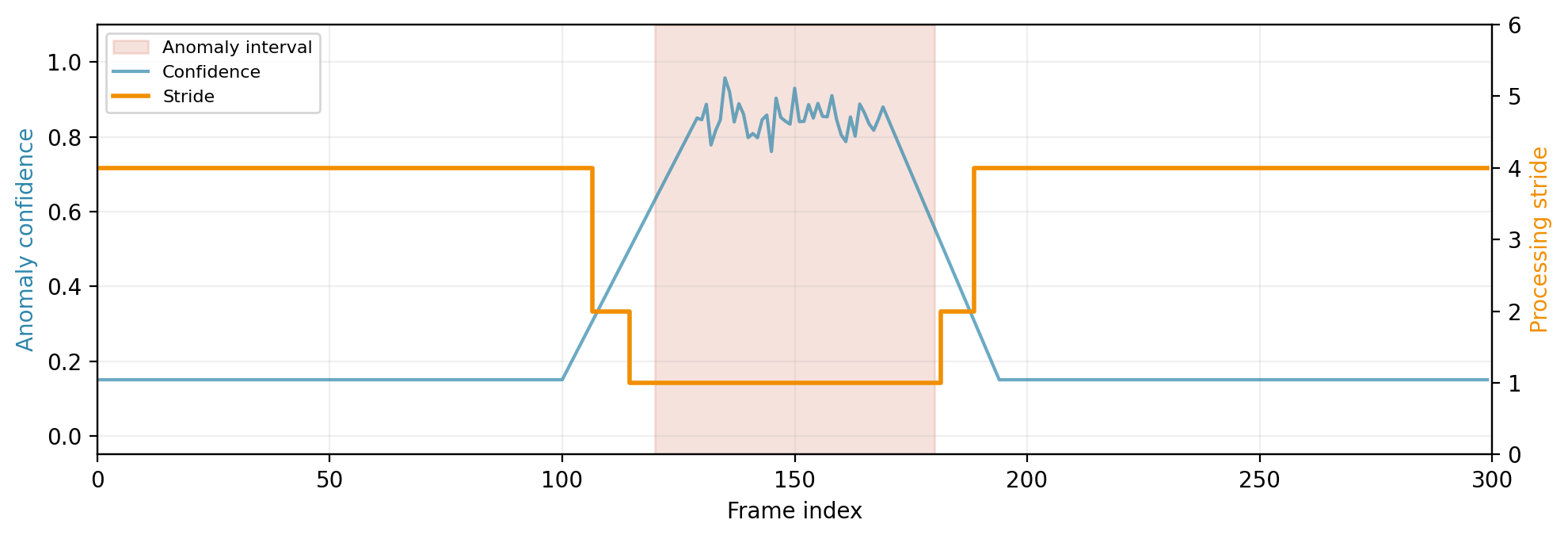}
  \caption{Adaptive stride over time for a sample UCF-Crime video. Red regions indicate anomaly intervals. The ACS reduces stride (processes more frames) near anomaly boundaries and increases stride during normal segments.}
  \label{fig:ablation_acs}
\end{figure}

We compare the ACS against fixed-stride baselines and a random-skip baseline. Table~\ref{tab:acs_modes} shows that fixed stride-2 reduces FPS cost but drops AUC by 0.89 points relative to no skipping. Fixed stride-4 is faster but loses 2.10 points. Random skipping (average stride 2.5) loses 1.43 points. The ACS achieves higher throughput than stride-2 while matching or slightly exceeding the no-skip AUC, because it allocates the saved compute to uncertain intervals where re-evaluation improves predictions. Figure~\ref{fig:ablation_acs} visualizes the adaptive stride over time for a sample video: the scheduler processes every frame near anomaly boundaries and skips frames during long normal intervals.

\begin{table}[t]
  \centering
  \caption{Comparison of frame skipping strategies on UCF-Crime.}
  \label{tab:acs_modes}
  \small
  \begin{tabular}{lcc}
    \toprule
    Strategy & AUC (\%) & FPS \\
    \midrule
    No skipping (stride 1)   & 87.81 & 33.4 \\
    Fixed stride 2            & 86.92 & 38.6 \\
    Fixed stride 4            & 85.71 & 43.1 \\
    Random skip (avg.\ 2.5)   & 86.38 & 40.2 \\
    \textbf{ACS (ours)}       & \textbf{87.83} & \textbf{41.5} \\
    \bottomrule
  \end{tabular}
\end{table}

We also vary the ACS hyperparameters $\theta_{\text{low}}$, $\theta_{\text{anom}}$, and $S_{\max}$. Setting $\theta_{\text{low}} = 0.3$, $\theta_{\text{anom}} = 0.5$, and $S_{\max} = 4$ gives the best AUC--FPS trade-off. Increasing $S_{\max}$ to 8 raises FPS to 44.2 but drops AUC to 87.52\%. Decreasing $\theta_{\text{low}}$ to 0.2 makes the scheduler more conservative, yielding 87.84\% AUC at 39.8 FPS.

\subsection{Temporal Window Length}

The number of snippets $N$ per video during training affects both performance and memory. We train with $N \in \{100, 150, 200, 250, 300\}$. Performance increases from $N = 100$ (86.14\%) to $N = 200$ (87.83\%) and saturates at $N = 250$ (87.87\%). At $N = 300$ the model achieves 87.91\% but requires 40\% more GPU memory. We use $N = 200$ as the default.

\subsection{Cross-Dataset Generalization}

To test generalization, we train on UCF-Crime and evaluate directly on ShanghaiTech (without fine-tuning). This zero-shot transfer achieves 88.35\% AUC on ShanghaiTech, compared to 97.21\% when trained on ShanghaiTech itself. Among the baselines tested under the same protocol, RTFM obtains 84.72\% and UR-DMU obtains 86.91\%. The 1.44-point improvement of \textbf{VigilFormer} over UR-DMU suggests that the deformable attention mechanism learns more transferable spatial features than fixed-topology alternatives. Training on ShanghaiTech and testing on UCF-Crime yields 78.63\% AUC (vs.\ 75.40\% for RTFM and 77.12\% for UR-DMU), consistent with the same trend.

\begin{table}[t]
  \centering
  \caption{Cross-dataset generalization (AUC \%). Models are trained on one dataset and evaluated on the other without fine-tuning.}
  \label{tab:cross}
  \small
  \begin{tabular}{lcc}
    \toprule
    Method & UCF $\rightarrow$ ST & ST $\rightarrow$ UCF \\
    \midrule
    RTFM~\cite{tian2021rtfm}    & 84.72 & 75.40 \\
    UR-DMU~\cite{zhou2023urdmu} & 86.91 & 77.12 \\
    \textbf{VigilFormer} (ours) & \textbf{88.35} & \textbf{78.63} \\
    \bottomrule
  \end{tabular}
\end{table}

% ============================================================
% 6  CONCLUSION
% ============================================================
\section{Conclusion}
\label{sec:conclusion}

We have presented \textbf{VigilFormer}, a framework for real-time weakly-supervised video anomaly detection that integrates deformable spatio-temporal attention, causal temporal convolutions with a contrastive MIL loss, and an adaptive confidence-based frame scheduler. The deformable encoder captures irregular spatial patterns at linear cost, the causal classifier produces temporally coherent anomaly scores trainable with only video-level labels, and the adaptive scheduler reallocates compute from static normal scenes to ambiguous or anomalous intervals. Experiments on three standard benchmarks show that \textbf{VigilFormer} achieves competitive or superior AUC while running at 41.5 FPS on a single GPU, addressing a practical gap between accuracy and throughput in surveillance deployments.

Limitations remain. The method relies on pre-extracted I3D features, which constrains the spatial resolution of the deformable attention and prevents end-to-end training from raw pixels. Per-category analysis reveals that subtle anomalies such as shoplifting and stealing still lag behind motion-salient categories. Future work will explore joint backbone fine-tuning with mixed-precision training to enable pixel-level deformable attention, and will investigate incorporating audio or text cues to improve detection of visually subtle events.

\bibliographystyle{plain}
\bibliography{references}

\end{document}